%% file: arxiv.tex
\documentclass{article}

\usepackage[preprint,nonatbib]{neurips_2024}
\usepackage{amsmath}

\usepackage[utf8]{inputenc} 
\usepackage[T1]{fontenc}    
\usepackage{hyperref}       
\usepackage{url}            
\usepackage{booktabs}       
\usepackage{amsfonts}       
\usepackage{nicefrac}       
\usepackage{microtype}      
\usepackage{xcolor}         
\usepackage{graphicx}
\usepackage{enumitem}
\newcommand{\eg}{\textit{e.g.}, }
\newcommand{\ie}{\textit{i.e.}, }
\usepackage{hyperref}
\usepackage{authblk}

\title{FreeLong: Training-Free Long Video Generation with SpectralBlend Temporal Attention}

\author[1]{\textbf{Yu Lu}}
\author[1]{\textbf{Yuanzhi Liang}}
\author[2]{\textbf{Linchao Zhu}}
\author[2]{\textbf{Yi Yang}}

\affil[ ]{\textit{ReLER Lab}}
\affil[1]{University of Technology Sydney}
\affil[2]{Zhejiang University}
\affil[ ]{\texttt{aniki.yulu@gmail.com,liangyzh18@outlook.com}}
\affil[ ]{\texttt{zhulinchao7@gmail.com,yangyics@zju.edu.cn}}

\begin{document}

\maketitle

\begin{abstract}

Video diffusion models have made substantial progress in various video generation applications. However, training models for long video generation tasks require significant computational and data resources, posing a challenge to developing long video diffusion models.
This paper investigates a straightforward and training-free approach to extend an existing short video diffusion model (\eg pre-trained on 16-frame videos) for consistent long video generation (\eg 128 frames). 
Our preliminary observation has found that directly applying the short video diffusion model to generate long videos can lead to severe video quality degradation. Further investigation reveals that this degradation is primarily due to the distortion of high-frequency components in long videos, characterized by a decrease in spatial high-frequency components and an increase in temporal high-frequency components. Motivated by this, we propose a novel solution named FreeLong to balance the frequency distribution of long video features during the denoising process.
FreeLong blends the low-frequency components of global video features, which encapsulate the entire video sequence, with the high-frequency components of local video features that focus on shorter subsequences of frames. This approach maintains global consistency while incorporating diverse and high-quality spatiotemporal details from local videos, enhancing both the consistency and fidelity of long video generation.
We evaluated FreeLong on multiple base video diffusion models and observed significant improvements. Additionally, our method supports coherent multi-prompt generation, ensuring both visual coherence and seamless transitions between scenes. Project page: 
\href{https://freelongvideo.github.io/}{https://yulu.net.cn/freelong}

\end{abstract}

\section{Introduction}

\begin{figure*}
\centering
   \setlength{\abovecaptionskip}{0.5cm}
   \includegraphics[scale = 0.37]{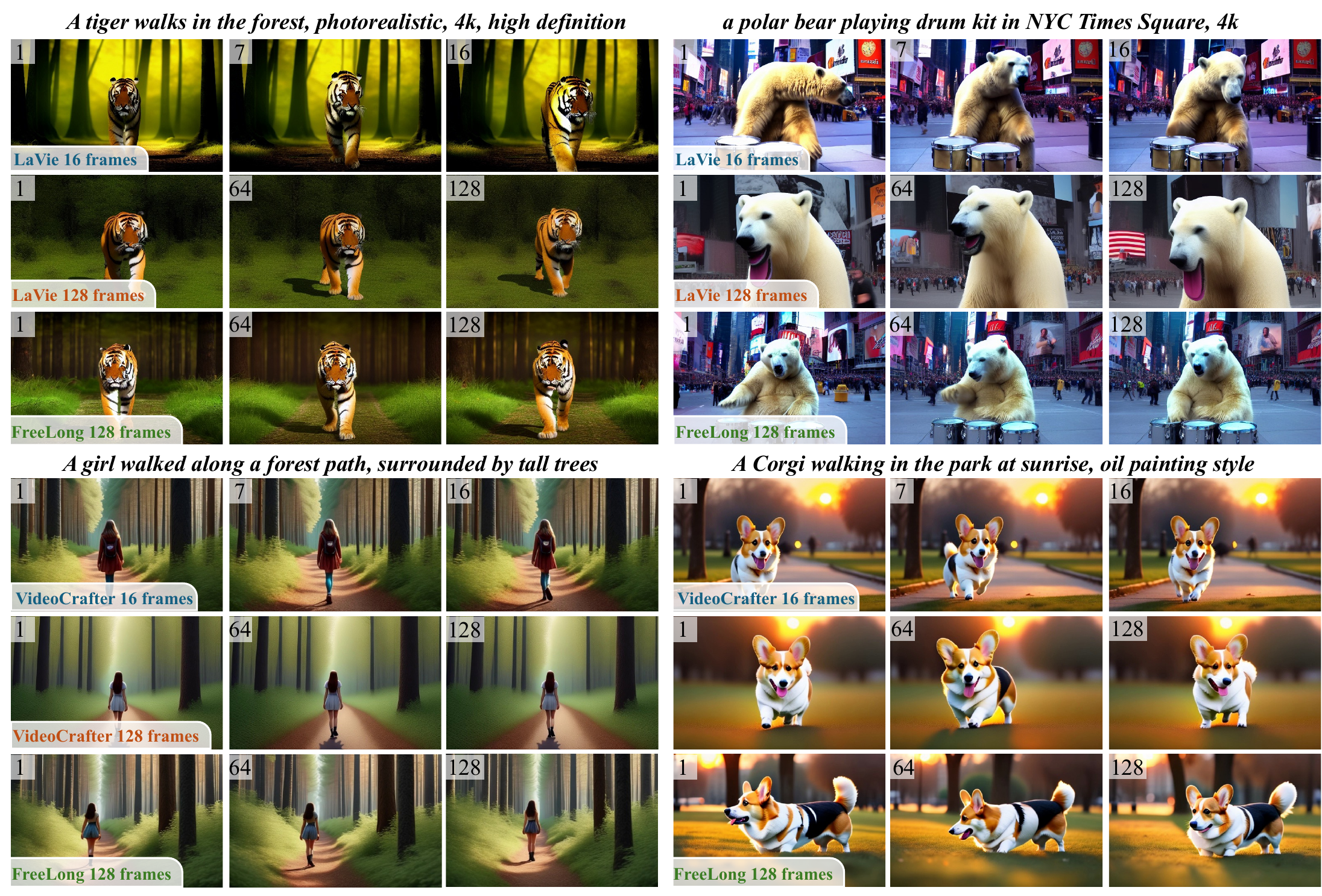}
\caption{
     \textbf{Results of Short and Long Videos.} The first row of each case shows 16-frame videos generated using short video diffusion models (LaVie~\cite{lavie} and VideoCrafter2~\cite{videocrafter}). Directly extending these models to longer videos, like those with 128 frames, preserves temporal consistency but lacks fine spatial-temporal details. In contrast, our proposed FreeLong adapts short video diffusion models to create consistent long videos with high fidelity.
}
   \label{fig:analysis_comparison}
\end{figure*}

Video diffusion models~\cite{lavie, videocrafter, cogvideo, animatediff, modelscope, flowzero,yang2024eva} trained on vast video-text datasets~\cite{internvid, webvid} have demonstrated impressive capabilities in generating high-quality videos. Inspired by Sora~\cite{sora}, multiple studies~\cite{streamt2v,vidu,keling} have concentrated on training these models to create longer videos using extensive, long video-text datasets~\cite{panda70m, hd130m, vlogger, videodrafter, lu2023show}. However, these methods demand significant computational resources and data annotations.

A more practical approach involves adapting pre-trained short video models to generate consistent longer video sequences without retraining. Recent research~\cite {freenoise, genlvideo} has explored sliding window temporal attention to ensure smooth transitions between video clips in the generation of long videos. Nonetheless, these techniques often struggle to maintain global temporal consistency across extended sequences and require multiple passes of temporal attention.

In this study, we propose a simple, training-free method to adapt existing short video diffusion models (e.g., pretrained on 16 frames) for generating consistent long videos (e.g., 128 frames). Initially, we examine the direct application of short video diffusion models for long video generation. As depicted in Figure~\ref{fig:analysis_comparison}, straightforwardly using a 16-frame video diffusion model to produce 128-frame sequences yields globally consistent yet low-quality results.

To delve further into these issues, we conducted a frequency analysis of the generated long videos. As shown in Figure~\ref{fig:frequency} (a), the low-frequency components remain stable as the video length increases, while the high-frequency components exhibit a noticeable decline, leading to a drop in video quality. The findings indicate that although the overall temporal structure is preserved, fine-grained details suffer notably in longer sequences. Specifically, there is a decrease in high-frequency spatial components (Figure~\ref{fig:frequency} (b)) and an increase in high-frequency temporal components (Figure~\ref{fig:frequency} (c)). This high-frequency distortion poses a challenge in maintaining high fidelity over extended sequences. As illustrated in the middle row of each case in Figure~\ref{fig:analysis_comparison}, intricate textures like forest paths or sunrises become blurred and less defined, while temporal flickering and sudden changes disrupt the video's narrative flow.

To tackle these challenges, we introduce FreeLong, a novel framework that employs SpectralBlend Temporal Attention (SpectralBlend-TA) to balance the frequency distribution of long video features in the denoising process. SpectralBlend-TA integrates global and local features via two parallel streams, enhancing the fidelity and consistency of long video generation. The global stream deals with the entire video sequence, capturing extensive dependencies and themes for narrative continuity. Meanwhile, the local stream focuses on shorter frame subsequences to retain fine details and smooth transitions, preserving high-frequency spatial and temporal information. SpectralBlend-TA combines global and local video features in the frequency domain, improving both consistency and fidelity by blending low-frequency global components with high-frequency local components. Our method is entirely training-free and allows for the easy integration of FreeLong into existing video diffusion models by adjusting the original temporal attention of video diffusion models. Comparative experiments demonstrate significant improvements in temporal consistency and video fidelity when applying our method to generate long video sequences.

Our contributions can be summarized as follows: 
\textbf{1)} We conduct a frequency analysis on the direct application of short video models for longer video generation and identify high-frequency distortions in the longer videos. 
\textbf{2)} We devise a SpectralBlend Temporal Attention mechanism to merge the consistent low-frequency components of global videos with the high-fidelity high-frequency components of local videos. 
\textbf{3)} Our training-free approach, FreeLong, outperforms existing state-of-the-art models in both temporal consistency and video fidelity.

\section{Related Work}
\input{sections/rela}

\input{sections/method}

\section{Experiments}

\subsection{Implementation Details}
\noindent\textbf{Baseline Models:} To evaluate the effectiveness and generalization of our proposed method, we apply FreeLong on two publicly available diffusion-based text-to-video models, LaVie~\cite{lavie} and VideoCrafter~\cite{videocrafter}. These models are trained on short videos with fixed length~(\ie 16 frames), we extend them to produce long videos (\ie 128 frames~\cite{survey}). We set $\alpha=8$ for the local attention setting and set $\tau$ to 25.
During inference, the parameters of the frequency filter for each model are kept the same for a fair comparison. Specifically, we use a Gaussian Low Pass Filter (GLPF) $\mathcal{P}_G$ with a normalized spatiotemporal stop frequency of $D_0 = 0.25$. Multi-prompt videos are generated with random noise, and FreeNoise~\cite{freenoise} is used for single-prompt long video generation.

\textbf{Test Prompts:} We chose 200 prompts from VBench~\cite{vbench} to validate the effectiveness of our method. 

\noindent\textbf{Evaluation Metrics:} For text-to-video generation, we employed several metrics from VBench~\cite{vbench} to evaluate two aspects: video consistency and video fidelity.
For video consistency measurement, we use two metrics: 
1). Subject consistency, computed by the DINO~\cite{dino} feature similarity across frames to assess whether object appearance remains consistent throughout the whole video. 
2). Background consistency, calculated by CLIP~\cite{clip} feature similarity across frames.
For video fidelity measurement, we use three metrics:
1). Motion smoothness, which utilizes the motion priors in the video frame interpolation model AMT~\cite{amt} to evaluate the smoothness of generated motions.
2). Temporal flickering, which takes static frames and computes the mean absolute difference across frames.
3). Imaging quality, calculated using the MUSIQ~\cite{musiq} image quality predictor trained on the SPAQ~\cite{spaq} dataset.

\subsection{Quantitative Comparison}

\begin{table}[t]
\centering
\small
\caption{\textbf{Quantitative Comparison}. ``Direct sampling'' and ``Sliding window'' indicate directly sampling 128 frames and applying temporal sliding windows based on short video generation models, respectively. Compared to these methods, our FreeLong achieves consistent long video generation with high fidelity.}
\begin{tabular}{l|ccccc|c}
\toprule
Methods & Sub ($\uparrow$) & Back ($\uparrow$) & Motion ($\uparrow$) & Flicker ($\uparrow$) & Imaging ($\uparrow$) & Inference Time ($\downarrow$)\\
\midrule
Direct sampling & 88.95 & 93.23 & 92.77 & 91.44 & 64.76  & 1.8s \\
Sliding window& 85.80 & 92.83 & 95.79 & 94.00 & 66.57 &  2.6s  \\
FreeNoise~\cite{freenoise} & 92.30 & 95.87 & 96.32 & 94.94 & 67.14 &  2.6s  \\
\midrule
\textit{Ours} & \textbf{95.16} & \textbf{96.80} & \textbf{96.85} & \textbf{96.04} & \textbf{67.55}  & \textbf{2.2s}  \\
\bottomrule
\end{tabular}

\label{tab:comapre_lavie}
\end{table}

\begin{figure*}
\centering
   \setlength{\abovecaptionskip}{0.5cm}
   \includegraphics[scale = 0.29]{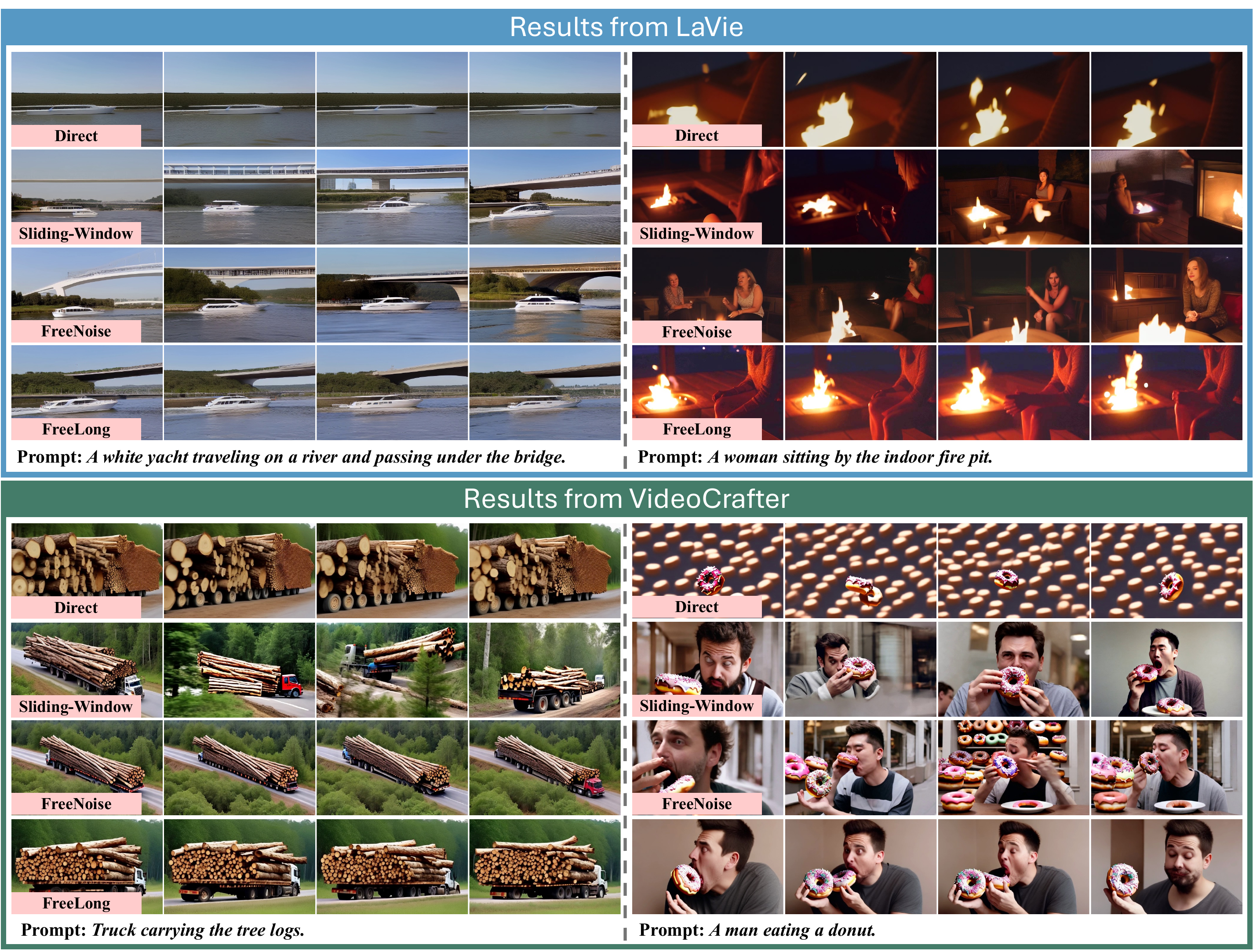}
     \caption{
\textbf{Qualitative Comparison.} Results from LaVie~\cite{lavie} and VideoCrafter~\cite{videocrafter} are presented. Direct videos exhibit consistent frames, but they appear over-smoothed. FreeNoise and the sliding-window approach struggle to capture global consistency effectively.
Our FreeLong method achieves consistent long video generation while maintaining high fidelity, preserving crucial details and textures across the entire sequence.
         }
   \label{fig:comparison}
\end{figure*}

We compare our FreeLong method with other training-free approaches for long video generation using diffusion models. Our comparison includes three methods: 
(1) Direct sampling. It directly samples 128 frames from the short video models. (2) Sliding window. It adopts temporal sliding windows~\cite{genlvideo} to process a fixed number of frames at a time. (3) FreeNoise~\cite{freenoise}. FreeNoise introduces repeat input noise to maintain temporal coherence across long sequences. 

Table~\ref{tab:comapre_lavie} presents the quantitative results. Direct generation of long videos suffers from high-frequency distortion, leading to significant quality degradation. 
This method results in low fidelity scores, including imaging quality, temporal flickering, and motion smoothness. The sliding-window method and FreeNoise show improved video quality thanks to the fixed effective temporal attention window but still face challenges in maintaining consistency across long videos. Our FreeLong method achieves the highest scores across all metrics, producing consistent long videos with high fidelity.
Moreover, we also examine the inference time of these methods on the NVIDIA A100. As delineated in Table~\ref{tab:comapre_lavie}, our approach achieves a faster speed compared to preceding methods by employing single-pass temporal attentions.

\subsection{Qualitative Comparison}
The synthesis results of each method are shown in Figure~\ref{fig:comparison}. In the first row, directly sampling 128 frames through a model trained on 16 frames will bring poor quality results due to the high-frequency distortion. For example, the yacht~(left) and the girl~(right) have blurred and the background is not clear. As shown in second row in Figure~\ref{fig:comparison}, using temporal sliding windows helps generate more vivid videos, but this approach ignores long-range visual consistency, causing the subject and background to appear significantly different across frames. 
FreeNoise attempts to promote global consistency by repeating and shuffling the initial noise for each frame; however, it fails to maintain long-range visual consistency and suffers from content mutations.
In contrast, our method, FreeLong, explicitly enforces global constraints during the denoising process, achieving temporal consistency while preserving high fidelity across frames. Results shown in Figure~\ref{fig:comparison} demonstrate that FreeLong successfully renders temporal consistent longer videos, outperforming all other methods.

\begin{figure*}
\centering
   \setlength{\abovecaptionskip}{0.5cm}
   \includegraphics[scale = 0.29]{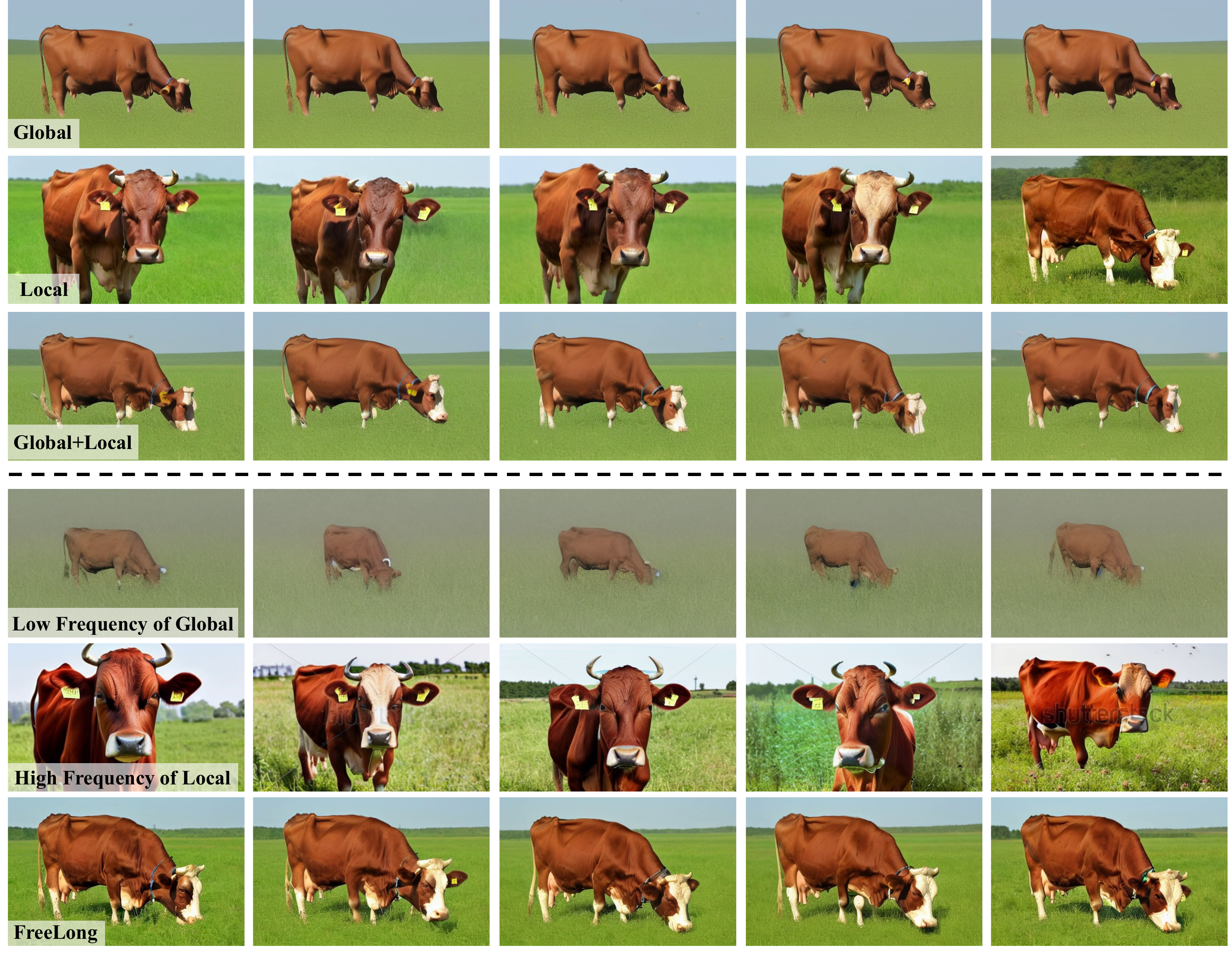}
     \caption{
\textbf{Ablation Study.} Global features and low-frequency components of global features ensure consistency but degrade fidelity. Local features and high-frequency local features maintain spatial-temporal details but lack temporal consistency. Directly adding global and local features degrades fidelity. Our method achieves both high fidelity and temporal consistency.}
   \label{fig:ablation}
\end{figure*}
\subsection{Ablation Studies}

To validate the effectiveness of each module in our FreeLong method—global video feature, local video feature, and our combined approach—we present the generated results by ablating each component. 

As shown in the top part of Figure~\ref{fig:ablation}, videos generated solely from global video features maintain consistent content but suffer from severe fidelity degradation. Conversely, videos generated using only local video features preserve fidelity due to the fixed effective temporal attention window but fail to maintain temporal consistency, as evidenced by the changing color of the cow. Simply combining global and local video features results in fidelity degradation because the high-frequency components of the global video features degrade significantly.

In the bottom part of Figure~\ref{fig:ablation}, we show the videos generated by combining the low-frequency components from global video features with the high-frequency components from local video features. Our approach effectively combines the consistency of global videos with the high fidelity of local videos, achieving both high fidelity and temporal consistency.

\subsection{Multi-Prompt Video Generation}

\begin{figure*}
\centering
   \setlength{\abovecaptionskip}{0.5cm}
   \includegraphics[scale = 0.40]{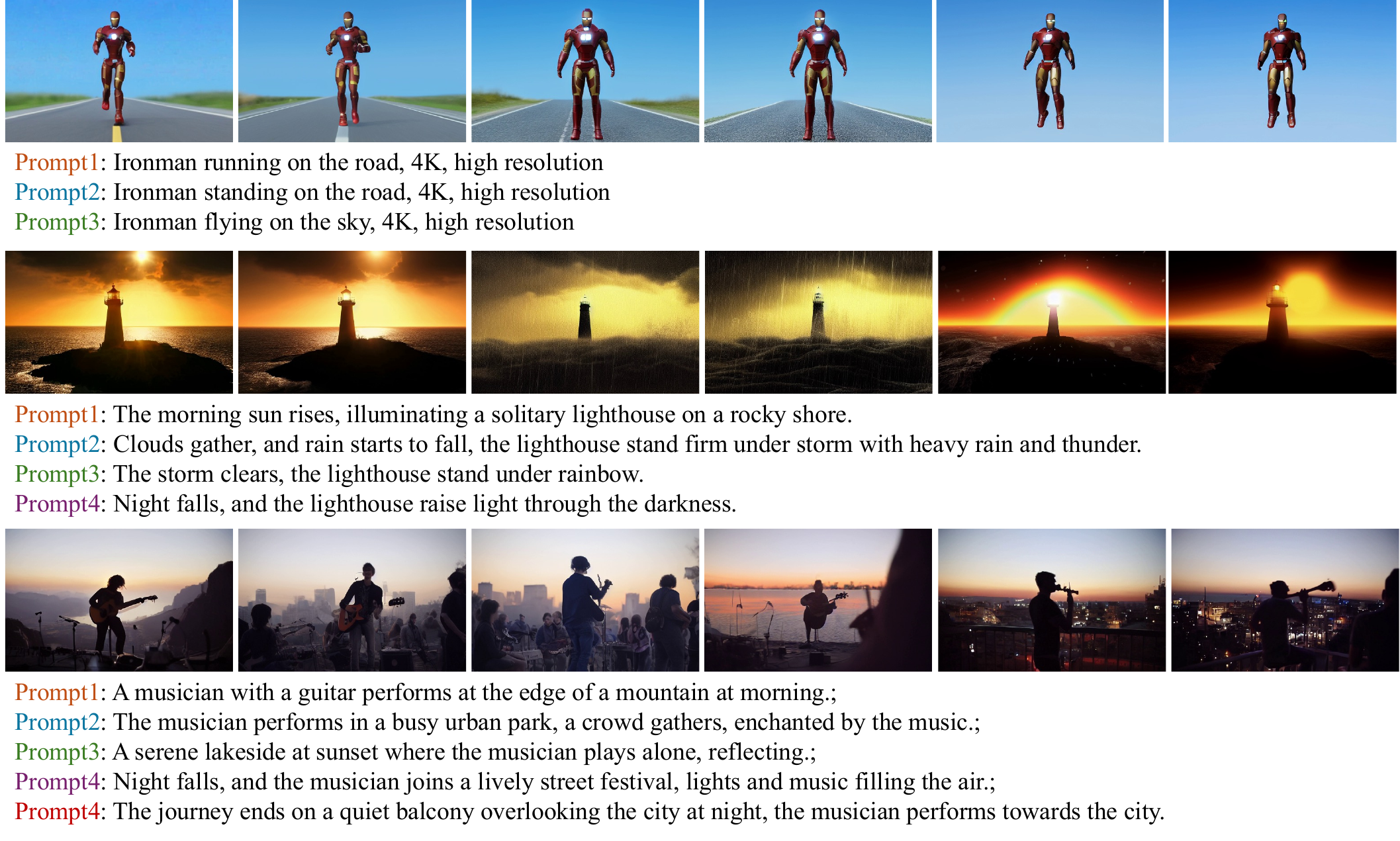}
   \vspace{-.2in}
     \caption{
\textbf{Results of Multi-Prompt Video Generation.} Our method ensures coherent visual continuity and motion consistency across different video segments.
         }
   \label{fig:multiprompt}
\end{figure*}

Our method can be seamlessly extended to multi-prompt video generation by providing different prompts for each video segment. As illustrated in Figure~\ref{fig:multiprompt}, our approach ensures coherent visual continuity and motion consistency. For instance, Ironman is shown running on the road, then standing, and finally flying into the sky, all within a consistent scene and with smooth action transitions.
In the second row, we demonstrate a more complex prompt sequence describing weather and scene transitions. Our method effectively models the transition from ``sunrise" to ``storm with heavy rain and thunder" to the final ``rainbow," maintaining consistency and capturing the fine-grained details of each prompt transition.

\section{Conclusion}
In this paper, we introduced FreeLong, a training-free method to adapt short video diffusion models for long video generation. 
Our research reveals that directly generating long videos from short video diffusion models results in poor quality, primarily due to high-frequency distortion. To resolve this issue, we employ the SpectralBlend Temporal Attention (SpectralBlend-TA) mechanism, which blends low-frequency global features with high-frequency local features to enhance consistency and fidelity in long videos. Our experiments demonstrate that FreeLong significantly outperforms existing models, achieving superior temporal consistency and video fidelity. 
Our experiments show that FreeLong significantly outperforms existing models, achieving better temporal consistency and video fidelity. FreeLong also supports coherent multi-prompt generation, offering a practical solution for high-quality long video creation without extensive retraining.

\bibliographystyle{unsrt}
\bibliography{ref}

\end{document}

%% file: sections/rela.tex

\textbf{Text-to-Video Diffusion Models:}
Text-to-video (T2V) generation has progressed significantly from early variational autoencoders~\cite{early1,early2} and GANs~\cite{gan} to advanced diffusion-based techniques~\cite{cogvideo,animatediff,MagicVideo,makeavideo,wu2023freeinit}, marking a major leap in synthesis methods. Modern video diffusion models build on pre-trained image-to-text diffusion models~\cite{sd,gligen,imagen}, incorporating temporal transformers in the diffusion UNet to capture temporal relationships. These models achieve impressive video generation results through post-training on video-text data~\cite{panda70m,webvid,internvid,lavie}, enhancing coherence and fidelity. However, due to computational constraints and limited dataset availability, current video diffusion models are typically trained on fixed-length short videos (\eg 16 frames), limiting their ability to produce longer videos. 
In this paper, we propose extending these short video diffusion models to generate long and consistent videos without requiring any additional training videos.

\textbf{Long-video Generation:}
Generating long videos is challenging due to temporal complexity, resource constraints, and the need for content consistency. Recent advancements focus on improving temporal coherence and visual quality using GAN-based~\cite{skorokhodov2022stylegan,brooks2022generating} and diffusion-based techniques~\cite{harvey2022flexible,voleti2022mcvd,tseng2023consistent,ho2020denoising,nuwa}. For instance, Nuwa-XL~\cite{nuwa} employs a parallel diffusion process, while StreamingT2V~\cite{streamt2v} uses an autoregressive approach with a short-long memory block to improve the consistency of long video sequences. Despite their effectiveness, these methods require substantial computational resources and large-scale datasets.
Recent research has explored training-free adaptations using short video diffusion models for long video generation. Gen-L-Video~\cite{wang2023gen} extends videos by merging overlapping sub-segments with a sliding-window method during denoising. FreeNoise~\cite{freenoise} employs sliding-window temporal attention and a noise initialization strategy to maintain temporal consistency. However, these approaches focus on smooth transitions between video clips and fail to capture global consistency across long video sequences. This paper proposes FreeLong, a novel approach that blends global and local video features during the denoising process to enhance both global temporal consistency and visual quality in long video generation.

%% file: sections/method.tex
\section{Observation and Analysis}

When attempting to adapt short video diffusion models to generate long videos, a straightforward approach is to input a longer noise sequence into the short video models. The temporal transformer layers in the video diffusion model are not constrained by input length, making this method seemingly viable. However, our empirical study reveals significant challenges, as demonstrated in Figure~\ref{fig:analysis_comparison}. Generated long videos often exhibit fewer detailed textures, such as blurred forests in the background, and more irregular variations, like abrupt changes in motion. We attribute these issues to two main factors: the limitations of the temporal attention mechanism and the distortion of high-frequency components.

\begin{figure*}
\centering
   \setlength{\abovecaptionskip}{0.5cm}
   \includegraphics[scale = 0.55]{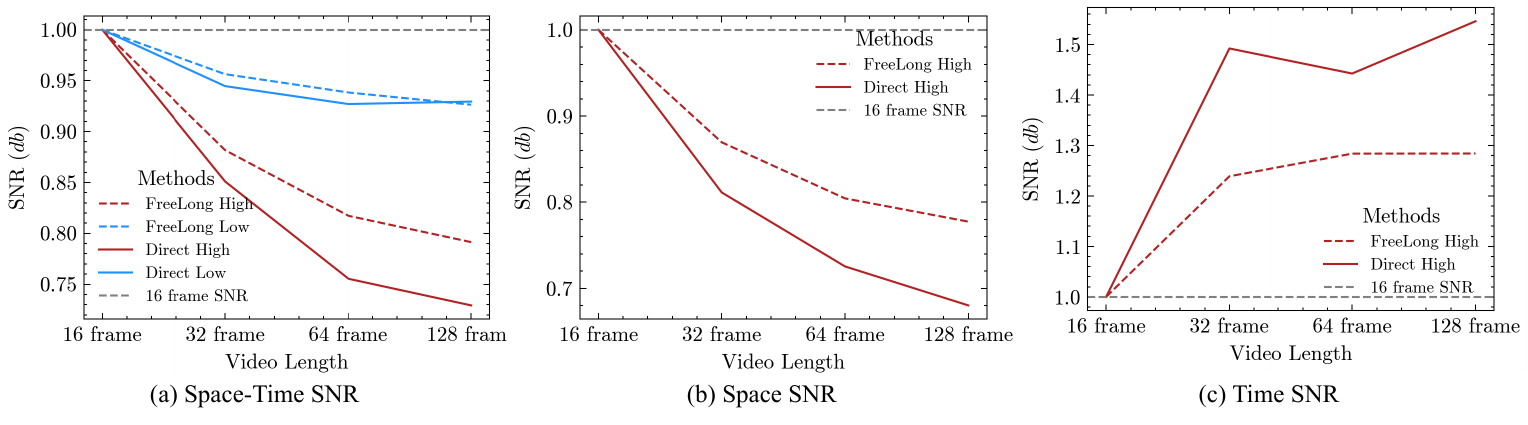}
     \caption{
     \textbf{Ratio of short video SNR on high/low frequency to different long videos.} 
    Our findings reveal that: (a) When direct extend short video diffusion model to generate long videos, the SNR of high-frequency components in the space-time frequency domain degrades significantly as video length increases. (b) In the spatial frequency domain, the SNR of high-frequency components decreases even more substantially, resulting in the over-smoothing of each frame. (c) Conversely, in the temporal frequency domain, the SNR of high-frequency components increases significantly, introducing temporal flickering.
              }
   \label{fig:frequency}
\end{figure*}

\begin{figure*}
\centering
   \setlength{\abovecaptionskip}{0.5cm}
   \includegraphics[scale = 0.34]{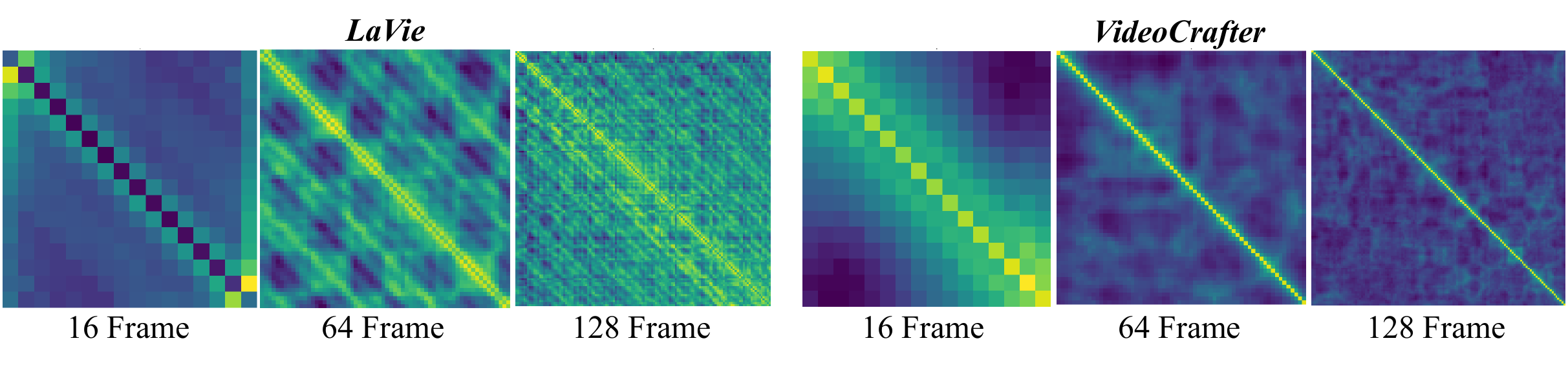}
     \caption{
     \textbf{Temporal Attention Visualization.} We visualize the temporal attention by average across all layers and time steps from LaVie~\cite{lavie} and VideoCrafter~\cite{videocrafter}. The attention maps for 16-frame videos exhibit a diagonal-like pattern, indicating a high correlation with adjacent frames, which helps preserve high-frequency details and motion patterns when generating new frames. In contrast, attention maps for longer videos are less structured, such as 128 frames, making the model struggle to identify and attend to the relevant information across distant frames. This lack of structure in the attention maps results in the distortion of high-frequency components of long videos, which results in the degradation of fine spatial-temporal details.
         }
   \label{fig:attention}
\end{figure*}

\textbf{Attention Mechanism Analysis:} The temporal attention mechanism in video diffusion models is pre-trained on fixed-length videos, which complicates its ability to generate longer videos. As shown in Figure~\ref{fig:attention}
, increasing video length hinders the temporal attention's ability to accurately capture frame-to-frame relationships. For 16-frame videos, the attention maps show a diagonal pattern, indicating high correlations with adjacent frames that preserve spatial-temporal details and motion patterns. In contrast, for 128-frame videos, the less structured attention maps suggest difficulty in focusing on relevant information across distant frames, leading to missed subtle motion patterns and over-smoothed or blurred generations.

\textbf{Frequency Analysis:} To better understand the generation process of long videos, we analyzed the frequency components in videos of varying lengths using the Signal-to-Noise Ratio (SNR) as a metric. Ideally, short video diffusion models generate 16-frame videos with high quality, and robust longer videos derived from such models should exhibit consistent SNR values across all frequency components. However, Figure~\ref{fig:frequency}
 reveals significant differences in the SNR of high/low frequency components\footnote{We split the frequency components into high-frequency (\( \phi \sim (0.25\pi - 1.00\pi) \)) and low-frequency (\( \phi \sim (0.00\pi - 0.25\pi) \)) and compared the SNR of each component in long videos to the corresponding SNR in 16-frame videos.} between generated short and long videos. The SNR of low-frequency components remains relatively consistent for long videos (1.0 for 16 frames to 0.93 for 128 frames), suggesting that the model maintains overall structure and low-frequency details in extended sequences. However, the SNR of high-frequency components drops significantly for longer videos (1.0 for 16 frames to 0.73 for 128 frames), indicating a loss of fine details and increased distortion, leading to suboptimal visual fidelity.

Further investigation into the spatial and temporal frequency domains revealed two key findings: (1) In the spatial domain, the high-frequency components of long videos degrade significantly (0.68 for 128 frames), causing substantial degradation of spatial details in each frame and resulting in blurred frames. (2) In the temporal domain, the high-frequency components increase with video length (1.5 for 128-frame videos), resulting in temporal flickering and incoherent video outputs.

\section{FreeLong: Training-free Long Video Generation}

\begin{figure*}
\centering
   \setlength{\abovecaptionskip}{0.5cm}
   \includegraphics[scale = 0.47]{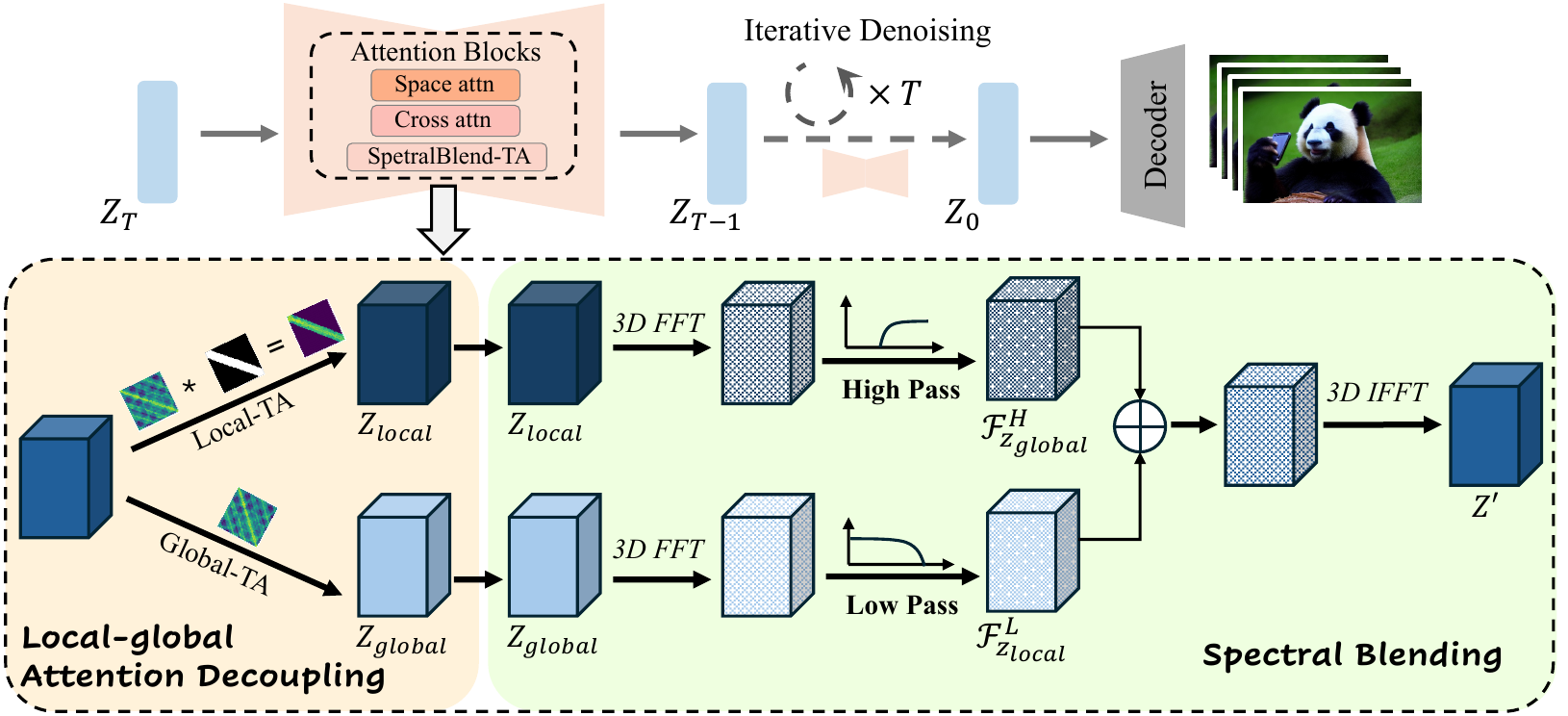}
     \caption{
    \textbf{Overview of FreeLong.} FreeLong facilitates consistent and high-fidelity video generation using SpectralBlend Temporal Attention (SpectralBlend-TA). SpectralBlend-TA effectively blends low-frequency global video features with high-frequency local video features through a two-step process: local-global attention decoupling and spectral blending. Local video features are obtained by masking temporal attention to concentrate on fixed-length adjacent frames, while global temporal attention encompasses all frames. During spectral blending, 3D FFT projects features into the frequency domain, where high-frequency local components and low-frequency global components are merged. The resulting blended feature, transformed back to the time domain via IFFT, is then utilized in the subsequent block for refined video generation.
             }
   \label{fig:pipeline}
\end{figure*}

Motivated by the above analysis, we propose FreeLong, a method designed to generate high-fidelity and consistent long videos using the inherent power of the diffusion model. As illustrated in Figure~\ref{fig:pipeline}, our FreeLong uses a diffusion UNet from pre-trained short video diffusion models and introduces a SpectralBlend Temporal Attention (SpectralBlend-TA) to facilitate long video generation. The SpectralBlend-TA consists of two steps: local-global attention decoupling and spectral blending.

\textbf{Local-global Attention Decoupling: }

The temporal attention in short video models is optimized to model short frame sequences accurately, maintaining high-fidelity visual information. 
Conversely, the long-range temporal attention from short video models tends to maintain overall layout and and object consistency.  
Given these properties, we first decouple the local and global attention. The local attention matrix can be obtained as:
\begin{equation}
A_{\text{local}}(i, j) = \begin{cases} 
\text{Softmax}\left(\frac{Q_i K_j^\top}{\sqrt{d}}\right) & \text{if } |i - j| \leq \alpha \\
0 & \text{otherwise}, 
\end{cases}
\end{equation}
where \( Q \) and \( K \) are the query and key matrices derived from the input video feature~\( Z_{in} \). The local attention \( A_{\text{local}} \) leads to each frame \( i \) only attending to frames within a window of \( 2\alpha \) frames. 
Given the local attention matrix \( A_{\text{local}} \), the local video features \( Z_{\text{local}} \) can be obtained by:
$Z_{\text{local}} = A_{\text{local}} V$, 
where \( V \) is the value matrix derived from the input video feature \( Z_{in} \).
By restricting the temporal attention to adjacent local frames, we preserve the capabilities of short video models, thereby retaining high-fidelity visual details in local video features.

We then define the global attention matrix where each frame attends to all other frames. The global attention matrix can be computed as follows:
\begin{equation}
A_{\text{global}}(i, j) = \text{Softmax}\left(\frac{Q_i K_j^\top}{\sqrt{d}}\right), 
\end{equation}
Given the global attention matrix \( A_{\text{global}} \), the global video features \( Z_{\text{global}} \) can be obtained by:
$Z_{\text{global}} = A_{\text{global}} V$. 
The global video features process the entire video sequence, ensuring narrative continuity and coherence, while capturing long-range dependencies and overarching themes.

\textbf{Spectral Blending:}  
After obtaining the global and local video features, a frequency filter is used to blend the low-frequency components of the global video latent \( Z_{global} \) with the high-frequency components of the local video latent \( Z_{local} \), resulting in a new video latent \( Z' \). This fused latent retains the global coherence and structure provided by \( Z_{global} \), while benefiting from the enhanced high-frequency details introduced by \( Z_{local} \). The process is described by:
\begin{align}
& \mathcal{F}^L_{z_{global}} = \text{FFT}_\text{3D}(Z_{global}) \odot \mathcal{P},  \\
& \mathcal{F}^H_{z_{local}} = \text{FFT}_\text{3D}(Z_{local}) \odot (1 - \mathcal{P}), \\
& Z' = \text{IFFT}_\text{3D}(\mathcal{F}^L_{z_{global}} + \mathcal{F}^H_{z_{local}})
\end{align}
where \(\text{FFT}_\text{3D}\) is the Fast Fourier Transformation operated on both spatial and temporal dimensions, \(\text{IFFT}_\text{3D}\) is the Inverse Fast Fourier Transformation that maps back the blended representation \( Z' \) from the frequency domain, and \(\mathcal{P} \in \mathbb{R}^{4 \times N \times h \times w}\) is the spatial-temporal Low Pass Filter (LPF), which is a tensor of the same shape as the latent. The final fused video feature \( Z' \) serves as the input to our subsequent video generation module. 

The rationale behind using low-frequency components from the global video features and high-frequency components from the local video features stems from our analysis. The global features provide a stable, coherent structure, preserving the overall layout and object consistency throughout the video. This is crucial for maintaining temporal consistency in long videos. On the other hand, local features retain high-fidelity details, which are essential for capturing fine textures and intricate motion patterns that tend to degrade in long sequences. By blending these components in the frequency domain, we harness the strengths of both global consistency and local detail preservation, addressing the issues of blurred frames and temporal flickering observed in our analysis.

Recent studies~\cite{ediff,masactrl} indicate that latent diffusion models~\cite{sd} generate varying levels of visual content at different stages of the denoising process: scene layout and object shapes in the early steps, and fine details in the later steps. We propose fusing global and local video features in the early \(\tau\) steps of the denoising process and using local video features in the remaining steps. This fusion ensures that the overall layout and object appearance of the generated long video follow the global features, thereby maintaining temporal consistency in the generated videos.

%% file: arxiv.bbl
\begin{thebibliography}{10}

\bibitem{lavie}
Yaohui Wang, Xinyuan Chen, Xin Ma, Shangchen Zhou, Ziqi Huang, Yi~Wang, Ceyuan Yang, Yinan He, Jiashuo Yu, Peiqing Yang, et~al.
\newblock Lavie: High-quality video generation with cascaded latent diffusion models.
\newblock {\em arXiv preprint arXiv:2309.15103}, 2023.

\bibitem{videocrafter}
Haoxin Chen, Yong Zhang, Xiaodong Cun, Menghan Xia, Xintao Wang, Chao Weng, and Ying Shan.
\newblock Videocrafter2: Overcoming data limitations for high-quality video diffusion models, 2024.

\bibitem{cogvideo}
Wenyi Hong, Ming Ding, Wendi Zheng, Xinghan Liu, and Jie Tang.
\newblock Cogvideo: Large-scale pretraining for text-to-video generation via transformers.
\newblock {\em arXiv preprint arXiv:2205.15868}, 2022.

\bibitem{animatediff}
Yuwei Guo, Ceyuan Yang, Anyi Rao, Yaohui Wang, Yu~Qiao, Dahua Lin, and Bo~Dai.
\newblock Animatediff: Animate your personalized text-to-image diffusion models without specific tuning.
\newblock {\em arXiv preprint arXiv:2307.04725}, 2023.

\bibitem{modelscope}
Jiuniu Wang, Hangjie Yuan, Dayou Chen, Yingya Zhang, Xiang Wang, and Shiwei Zhang.
\newblock Modelscope text-to-video technical report.
\newblock {\em arXiv preprint arXiv:2308.06571}, 2023.

\bibitem{flowzero}
Yu~Lu, Linchao Zhu, Hehe Fan, and Yi~Yang.
\newblock Flowzero: Zero-shot text-to-video synthesis with llm-driven dynamic scene syntax.
\newblock {\em arXiv preprint arXiv:2311.15813}, 2023.

\bibitem{yang2024eva}
Xiangpeng Yang, Linchao Zhu, Hehe Fan, and Yi~Yang.
\newblock Eva: Zero-shot accurate attributes and multi-object video editing.
\newblock {\em arXiv preprint arXiv:2403.16111}, 2024.

\bibitem{internvid}
Yi~Wang, Yinan He, Yizhuo Li, Kunchang Li, Jiashuo Yu, Xin Ma, Xinhao Li, Guo Chen, Xinyuan Chen, Yaohui Wang, et~al.
\newblock Internvid: A large-scale video-text dataset for multimodal understanding and generation.
\newblock In {\em The Twelfth International Conference on Learning Representations}, 2023.

\bibitem{webvid}
Max Bain, Arsha Nagrani, G{\"u}l Varol, and Andrew Zisserman.
\newblock Frozen in time: A joint video and image encoder for end-to-end retrieval.
\newblock In {\em IEEE International Conference on Computer Vision}, 2021.

\bibitem{sora}
Tim Brooks, Bill Peebles, Connor Holmes, Will DePue, Yufei Guo, Li~Jing, David Schnurr, Joe Taylor, Troy Luhman, Eric Luhman, Clarence Ng, Ricky Wang, and Aditya Ramesh.
\newblock Video generation models as world simulators, 2024.
\newblock Accessed: 2024-05-09.

\bibitem{streamt2v}
Roberto Henschel, Levon Khachatryan, Daniil Hayrapetyan, Hayk Poghosyan, Vahram Tadevosyan, Zhangyang Wang, Shant Navasardyan, and Humphrey Shi.
\newblock Streamingt2v: Consistent, dynamic, and extendable long video generation from text.
\newblock {\em arXiv preprint arXiv:2403.14773}, 2024.

\bibitem{vidu}
Fan Bao, Chendong Xiang, Gang Yue, Guande He, Hongzhou Zhu, Kaiwen Zheng, Min Zhao, Shilong Liu, Yaole Wang, and Jun Zhu.
\newblock Vidu: a highly consistent, dynamic and skilled text-to-video generator with diffusion models.
\newblock {\em arXiv preprint arXiv:2405.04233}, 2024.

\bibitem{keling}
Ye~Tian, Ling Yang, Haotian Yang, Yuan Gao, Yufan Deng, Jingmin Chen, Xintao Wang, Zhaochen Yu, Xin Tao, Pengfei Wan, et~al.
\newblock Videotetris: Towards compositional text-to-video generation.
\newblock {\em arXiv preprint arXiv:2406.04277}, 2024.

\bibitem{panda70m}
Tsai-Shien Chen, Aliaksandr Siarohin, Willi Menapace, Ekaterina Deyneka, Hsiang-wei Chao, Byung~Eun Jeon, Yuwei Fang, Hsin-Ying Lee, Jian Ren, Ming-Hsuan Yang, and Sergey Tulyakov.
\newblock Panda-70m: Captioning 70m videos with multiple cross-modality teachers.
\newblock {\em arXiv preprint arXiv:2402.19479}, 2024.

\bibitem{hd130m}
Wenjing Wang, Huan Yang, Zixi Tuo, Huiguo He, Junchen Zhu, Jianlong Fu, and Jiaying Liu.
\newblock Videofactory: Swap attention in spatiotemporal diffusions for text-to-video generation.
\newblock {\em arXiv preprint arXiv:2305.10874}, 2023.

\bibitem{vlogger}
Shaobin Zhuang, Kunchang Li, Xinyuan Chen, Yaohui Wang, Ziwei Liu, Yu~Qiao, and Yali Wang.
\newblock Vlogger: Make your dream a vlog.
\newblock {\em arXiv preprint arXiv:2401.09414}, 2024.

\bibitem{videodrafter}
Fuchen Long, Zhaofan Qiu, Ting Yao, and Tao Mei.
\newblock Videodrafter: Content-consistent multi-scene video generation with llm.
\newblock {\em arXiv preprint arXiv:2401.01256}, 2024.

\bibitem{lu2023show}
Yu~Lu, Feiyue Ni, Haofan Wang, Xiaofeng Guo, Linchao Zhu, Zongxin Yang, Ruihua Song, Lele Cheng, and Yi~Yang.
\newblock Show me a video: A large-scale narrated video dataset for coherent story illustration.
\newblock {\em IEEE Transactions on Multimedia}, 2023.

\bibitem{freenoise}
Haonan Qiu, Menghan Xia, Yong Zhang, Yingqing He, Xintao Wang, Ying Shan, and Ziwei Liu.
\newblock Freenoise: Tuning-free longer video diffusion via noise rescheduling.
\newblock {\em arXiv preprint arXiv:2310.15169}, 2023.

\bibitem{genlvideo}
Fu-Yun Wang, Wenshuo Chen, Guanglu Song, Han-Jia Ye, Yu~Liu, and Hongsheng Li.
\newblock Gen-l-video: Multi-text to long video generation via temporal co-denoising.
\newblock {\em arXiv preprint arXiv:2305.18264}, 2023.

\bibitem{early1}
Yitong Li, Martin~Renqiang Min, Dinghan Shen, David~E. Carlson, and Lawrence Carin.
\newblock Video generation from text.
\newblock {\em CoRR}, abs/1710.00421, 2017.

\bibitem{early2}
Yitong Li, Zhe Gan, Yelong Shen, Jingjing Liu, Yu~Cheng, Yuexin Wu, Lawrence Carin, David~E. Carlson, and Jianfeng Gao.
\newblock Storygan: {A} sequential conditional {GAN} for story visualization.
\newblock In {\em {IEEE} Conference on Computer Vision and Pattern Recognition, {CVPR} 2019, Long Beach, CA, USA, June 16-20, 2019}, pages 6329--6338. Computer Vision Foundation / {IEEE}, 2019.

\bibitem{gan}
Ian~J. Goodfellow, Jean Pouget{-}Abadie, Mehdi Mirza, Bing Xu, David Warde{-}Farley, Sherjil Ozair, Aaron~C. Courville, and Yoshua Bengio.
\newblock Generative adversarial networks.
\newblock {\em CoRR}, abs/1406.2661, 2014.

\bibitem{MagicVideo}
Daquan Zhou, Weimin Wang, Hanshu Yan, Weiwei Lv, Yizhe Zhu, and Jiashi Feng.
\newblock Magicvideo: Efficient video generation with latent diffusion models.
\newblock {\em CoRR}, abs/2211.11018, 2022.

\bibitem{makeavideo}
Uriel Singer, Adam Polyak, Thomas Hayes, Xi~Yin, Jie An, Songyang Zhang, Qiyuan Hu, Harry Yang, Oron Ashual, Oran Gafni, Devi Parikh, Sonal Gupta, and Yaniv Taigman.
\newblock Make-a-video: Text-to-video generation without text-video data.
\newblock In {\em The Eleventh International Conference on Learning Representations, {ICLR} 2023, Kigali, Rwanda, May 1-5, 2023}. OpenReview.net, 2023.

\bibitem{wu2023freeinit}
Tianxing Wu, Chenyang Si, Yuming Jiang, Ziqi Huang, and Ziwei Liu.
\newblock Freeinit: Bridging initialization gap in video diffusion models.
\newblock {\em arXiv preprint arXiv:2312.07537}, 2023.

\bibitem{sd}
Robin Rombach, Andreas Blattmann, Dominik Lorenz, Patrick Esser, and Bj{\"o}rn Ommer.
\newblock High-resolution image synthesis with latent diffusion models.
\newblock In {\em Proceedings of the IEEE/CVF conference on computer vision and pattern recognition}, pages 10684--10695, 2022.

\bibitem{gligen}
Yuheng Li, Haotian Liu, Qingyang Wu, Fangzhou Mu, Jianwei Yang, Jianfeng Gao, Chunyuan Li, and Yong~Jae Lee.
\newblock Gligen: Open-set grounded text-to-image generation.
\newblock In {\em Proceedings of the IEEE/CVF Conference on Computer Vision and Pattern Recognition}, pages 22511--22521, 2023.

\bibitem{imagen}
Chitwan Saharia, William Chan, Saurabh Saxena, Lala Li, Jay Whang, Emily~L Denton, Kamyar Ghasemipour, Raphael Gontijo~Lopes, Burcu Karagol~Ayan, Tim Salimans, et~al.
\newblock Photorealistic text-to-image diffusion models with deep language understanding.
\newblock {\em Advances in neural information processing systems}, 35:36479--36494, 2022.

\bibitem{skorokhodov2022stylegan}
Ivan Skorokhodov, Sergey Tulyakov, and Mohamed Elhoseiny.
\newblock Stylegan-v: A continuous video generator with the price, image quality and perks of stylegan2.
\newblock In {\em Proceedings of the IEEE/CVF Conference on Computer Vision and Pattern Recognition}, pages 3626--3636, 2022.

\bibitem{brooks2022generating}
Tim Brooks, Janne Hellsten, Miika Aittala, Ting-Chun Wang, Timo Aila, Jaakko Lehtinen, Ming-Yu Liu, Alexei Efros, and Tero Karras.
\newblock Generating long videos of dynamic scenes.
\newblock {\em Advances in Neural Information Processing Systems}, 35:31769--31781, 2022.

\bibitem{harvey2022flexible}
William Harvey, Saeid Naderiparizi, Vaden Masrani, Christian Weilbach, and Frank Wood.
\newblock Flexible diffusion modeling of long videos.
\newblock {\em Advances in Neural Information Processing Systems}, 35:27953--27965, 2022.

\bibitem{voleti2022mcvd}
Vikram Voleti, Alexia Jolicoeur-Martineau, and Chris Pal.
\newblock Mcvd-masked conditional video diffusion for prediction, generation, and interpolation.
\newblock {\em Advances in neural information processing systems}, 35:23371--23385, 2022.

\bibitem{tseng2023consistent}
Hung-Yu Tseng, Qinbo Li, Changil Kim, Suhib Alsisan, Jia-Bin Huang, and Johannes Kopf.
\newblock Consistent view synthesis with pose-guided diffusion models.
\newblock In {\em Proceedings of the IEEE/CVF Conference on Computer Vision and Pattern Recognition}, pages 16773--16783, 2023.

\bibitem{ho2020denoising}
Jonathan Ho, Ajay Jain, and Pieter Abbeel.
\newblock Denoising diffusion probabilistic models.
\newblock {\em Advances in neural information processing systems}, 33:6840--6851, 2020.

\bibitem{nuwa}
Shengming Yin, Chenfei Wu, Huan Yang, Jianfeng Wang, Xiaodong Wang, Minheng Ni, Zhengyuan Yang, Linjie Li, Shuguang Liu, Fan Yang, et~al.
\newblock Nuwa-xl: Diffusion over diffusion for extremely long video generation.
\newblock {\em arXiv preprint arXiv:2303.12346}, 2023.

\bibitem{wang2023gen}
Fu-Yun Wang, Wenshuo Chen, Guanglu Song, Han-Jia Ye, Yu~Liu, and Hongsheng Li.
\newblock Gen-l-video: Multi-text to long video generation via temporal co-denoising.
\newblock {\em arXiv preprint arXiv:2305.18264}, 2023.

\bibitem{ediff}
Yogesh Balaji, Seungjun Nah, Xun Huang, Arash Vahdat, Jiaming Song, Qinsheng Zhang, Karsten Kreis, Miika Aittala, Timo Aila, Samuli Laine, et~al.
\newblock ediff-i: Text-to-image diffusion models with an ensemble of expert denoisers.
\newblock {\em arXiv preprint arXiv:2211.01324}, 2022.

\bibitem{masactrl}
Mingdeng Cao, Xintao Wang, Zhongang Qi, Ying Shan, Xiaohu Qie, and Yinqiang Zheng.
\newblock Masactrl: Tuning-free mutual self-attention control for consistent image synthesis and editing.
\newblock In {\em Proceedings of the IEEE/CVF International Conference on Computer Vision}, pages 22560--22570, 2023.

\bibitem{survey}
Chengxuan Li, Di~Huang, Zeyu Lu, Yang Xiao, Qingqi Pei, and Lei Bai.
\newblock A survey on long video generation: Challenges, methods, and prospects.
\newblock {\em arXiv preprint arXiv:2403.16407}, 2024.

\bibitem{vbench}
Ziqi Huang, Yinan He, Jiashuo Yu, Fan Zhang, Chenyang Si, Yuming Jiang, Yuanhan Zhang, Tianxing Wu, Qingyang Jin, Nattapol Chanpaisit, et~al.
\newblock Vbench: Comprehensive benchmark suite for video generative models.
\newblock {\em arXiv preprint arXiv:2311.17982}, 2023.

\bibitem{dino}
Mathilde Caron, Hugo Touvron, Ishan Misra, Herv{\'e} J{\'e}gou, Julien Mairal, Piotr Bojanowski, and Armand Joulin.
\newblock Emerging properties in self-supervised vision transformers.
\newblock In {\em Proceedings of the IEEE/CVF international conference on computer vision}, pages 9650--9660, 2021.

\bibitem{clip}
Alec Radford, Jong~Wook Kim, Chris Hallacy, Aditya Ramesh, Gabriel Goh, Sandhini Agarwal, Girish Sastry, Amanda Askell, Pamela Mishkin, Jack Clark, et~al.
\newblock Learning transferable visual models from natural language supervision.
\newblock In {\em International conference on machine learning}, pages 8748--8763. PMLR, 2021.

\bibitem{amt}
Zhen Li, Zuo-Liang Zhu, Ling-Hao Han, Qibin Hou, Chun-Le Guo, and Ming-Ming Cheng.
\newblock Amt: All-pairs multi-field transforms for efficient frame interpolation.
\newblock In {\em Proceedings of the IEEE/CVF Conference on Computer Vision and Pattern Recognition}, pages 9801--9810, 2023.

\bibitem{musiq}
Junjie Ke, Qifei Wang, Yilin Wang, Peyman Milanfar, and Feng Yang.
\newblock Musiq: Multi-scale image quality transformer.
\newblock In {\em Proceedings of the IEEE/CVF international conference on computer vision}, pages 5148--5157, 2021.

\bibitem{spaq}
Yuming Fang, Hanwei Zhu, Yan Zeng, Kede Ma, and Zhou Wang.
\newblock Perceptual quality assessment of smartphone photography.
\newblock In {\em Proceedings of the IEEE/CVF conference on computer vision and pattern recognition}, pages 3677--3686, 2020.

\end{thebibliography}
